\documentclass[sigconf]{acmart}
\usepackage{graphicx}
\usepackage{subcaption}
\usepackage[linesnumbered,ruled,vlined]{algorithm2e}
\usepackage{multirow}
\usepackage{enumerate}
\usepackage{adjustbox}
\graphicspath{ {images/} }

\AtBeginDocument{%
  \providecommand\BibTeX{{%
    \normalfont B\kern-0.5em{\scshape i\kern-0.25em b}\kern-0.8em\TeX}}}

\begin{document}

\title{Towards Unsupervised Domain Adaptation for Deep Face Recognition under Privacy Constraints via Federated Learning}



\author{Weiming Zhuang$^{1, 2}$ \quad Xin Gan$^{1}$  \quad Yonggang Wen$^{1}$ \quad Xuesen Zhang$^{2}$ \quad Shuai Zhang$^{2}$ \quad Shuai Yi$^{2}$}
\affiliation{%
 $^{1}$Nanyang Technological University, Singapore \\ 
 $^{2}$SenseTime Research, China
 \country{}
}
\email{weiming001, ganx0005@e.ntu.edu.sg, ygwen@ntu.edu.sg}
\email{zhangxuesen, zhangshuai, yishuai@sensetime.com}

\renewcommand{\authors}{Weiming Zhuang, Xin Gan, Yonggang Wen, and Shuai Zhang}

\renewcommand{\shortauthors}{W. Zhuang et al.}

\settopmatter{printacmref=false}
\renewcommand\footnotetextcopyrightpermission[1]{} 
\pagestyle{plain}



\begin{abstract}





Unsupervised domain adaptation has been widely adopted to generalize models for unlabeled data in a target domain, given labeled data in a source domain, whose data distributions differ from the target domain. However, existing works are inapplicable to face recognition under privacy constraints because they require sharing sensitive face images between two domains. To address this problem, we propose a novel unsupervised federated face recognition approach (\textit{FedFR}). FedFR improves the performance in the target domain by iteratively aggregating knowledge from the source domain through federated learning. It protects data privacy by transferring models instead of raw data between domains. Besides, we propose a new domain constraint loss (\textit{DCL}) to regularize source domain training. DCL suppresses the data volume dominance of the source domain. We also enhance a hierarchical clustering algorithm to predict pseudo labels for the unlabeled target domain accurately. To this end, FedFR forms an end-to-end training pipeline: (1) pre-train in the source domain; (2) predict pseudo labels by clustering in the target domain; (3) conduct domain-constrained federated learning across two domains. Extensive experiments and analysis on two newly constructed benchmarks demonstrate the effectiveness of FedFR. It outperforms the baseline and classic methods in the target domain by over 4\% on the more realistic benchmark. We believe that FedFR will shed light on applying federated learning to more computer vision tasks under privacy constraints.

\end{abstract}




\maketitle

\begin{figure}[h]
\begin{center}
\includegraphics[width=\linewidth]{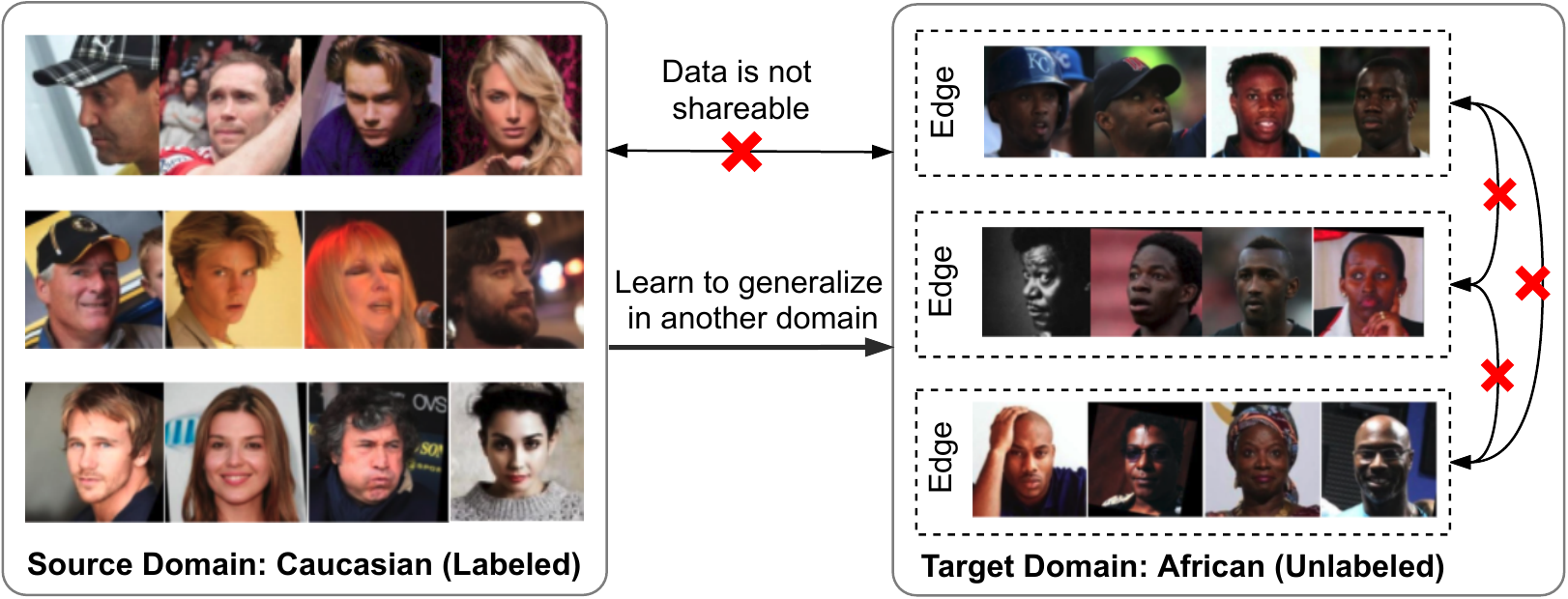}
\end{center}
    \caption{Illustration of unsupervised domain adaptation under privacy constraints. We aim to learn a model for the unlabeled target domain by adapting from the source domain without data sharing. Besides, unlabeled data are collected and stored in multiple edge devices in the target domain. Data is not shareable due to privacy protection regulations.}
\label{fig:problem}
\end{figure}

\section{Introduction}

Despite that face recognition using the deep neural network has achieved outstanding performances \cite{taigman2014deepface, sun2014deep-FR-10000, sun2014deep-FR-nips, schroff2015facenet, liu2017sphereface, deng2019arcface}, a well-trained model would fail to generalize across different attributes like age and ethnicity. For example, in cross-entity scenarios, a model trained in one region with face images of fair skin would not perform well in another with data of tan skin \cite{wang2019rfw, wang2020bupt}. It is known as the \textit{domain shift problem} --- the \textit{source domain} where models are trained share different distributions with the \textit{target domain} where models are deployed. Besides, the target domain data is mostly \textit{unlabeled} \cite{long2015dan}, so training a new model in the target domain is not feasible. 






Moreover, the domain shift problem is even more challenging in real-world scenarios where data is \textit{not shareable} between domains due to privacy constraint, as illustrated in Figure \ref{fig:problem}. The increasingly stringent regulations have limited data sharing between organizations in different countries, especially for sensitive data like face images \cite{gdpr}.  Besides, the source domain usually contains a large amount of data (e.g. high-resolution images), whose volume could be too large to transfer. In the target domain, data are unlabeled when collected and stored in multiple edge devices. Centralizing them would also imply potential risks of privacy leakage. Hence, navigating a solution to this challenging and realistic problem could have high practical values. 

Existing methods cannot fully solve the domain shift problem without data sharing. Collecting and labeling more data in the target domain is straightforward, but it is expensive and likely under scrutiny. Although unsupervised domain adaptation methods \cite{sohn2017uda-FR-video, luo2018uda-FR, sohn2018uda-metric-learning, wang2020uda-clustering-face, long2015dan, ganin2016dann, tzeng2017adda, long2018cada} reduce domain gaps effectively, they mostly assume data is shared between domains. Among these methods, model adaptation \cite{li2020model-adaptation} proposes to use only the target domain data on the image classification task, but it requires identical classes in both domains, which is impractical for face recognition because the identities (classes) of face images are different in domains.

Federated learning (\textit{FL}) is an emerging distributed machine learning solution to learn a model without data sharing among decentralized clients \cite{fedavg}. However, \textit{applying FL to unsupervised domain adaptation for face recognition is non-trivial}: (1) FL requires data labels, but the target domain is unlabeled. (2) FL equally weights the source and target domains, which is not optimal to deliver high performance in the target domain, especially when the source domain contains much more data. (3) FL requires identical models in clients, but the model structures of the loss vary in domains due to varied numbers of face identities. 

In this paper, we propose a novel unsupervised federated face recognition approach, \textit{FedFR}, to address unsupervised domain adaptation under privacy constraints. Firstly, for unlabeled data in the target domain, we enhance a hierarchical clustering algorithm \cite{sarfraz2019finch} by adding a distance constrain to accurately and efficiently generate pseudo labels. The pseudo labels are predicted by clustering features extracted from the unlabeled data using the model pre-trained in the source domain. Secondly, to tackle the domain shift problem, we propose to iteratively transfer knowledge from the source domain to the target domain via FL. Specifically, both domains conduct training with their data. A central server then aggregates their trained models and updates both domains with the aggregated global model. It protects data privacy by transmitting models instead of raw data. To this end, we form an end-to-end training pipeline: (1) source domain pre-training; (2) pseudo label generation; (3) FL across domains.

Besides, we introduce three enhancements on the FL algorithm: (1) As the source domain could contain much more data than the target domain, we propose a new domain constraint loss (DCL). DCL regularizes the model trained in the source domain not to deviate too far from the global model, such that the global model can lean towards the target domain; (2) FedFR aggregates only the backbone of face recognition model instead of the whole model; (3) FedFR uses local iterations as the minimum execution unit instead of local epochs to reduce training time.

We construct two new benchmarks to evaluate FedFR and conduct ablation studies to analyze it. Extensive empirical results show that FedFR effectively elevates the performance in the target domain, superior to the baselines and classic methods. It outperforms other methods by at least 4\% in the target domain on the more realistic benchmark. We also demonstrate that FedFR is communication-efficient and scalable in real-world scenarios.
In summary, our key contributions are: 
\begin{itemize}
    
    \item We investigate an important but overlooked problem --- unsupervised domain adaptation for face recognition under privacy constraints.

    
    \item We propose a novel unsupervised federated face recognition approach (FedFR) with an end-to-end training pipeline and multiple algorithm enhancements to solve the problem.
    
    \item We demonstrate that FedFR is effective and superior to the baseline and classic methods via extensive experiments on two newly constructed benchmarks.

\end{itemize}

\begin{figure*}[t]
\begin{center}
\includegraphics[width=\linewidth]{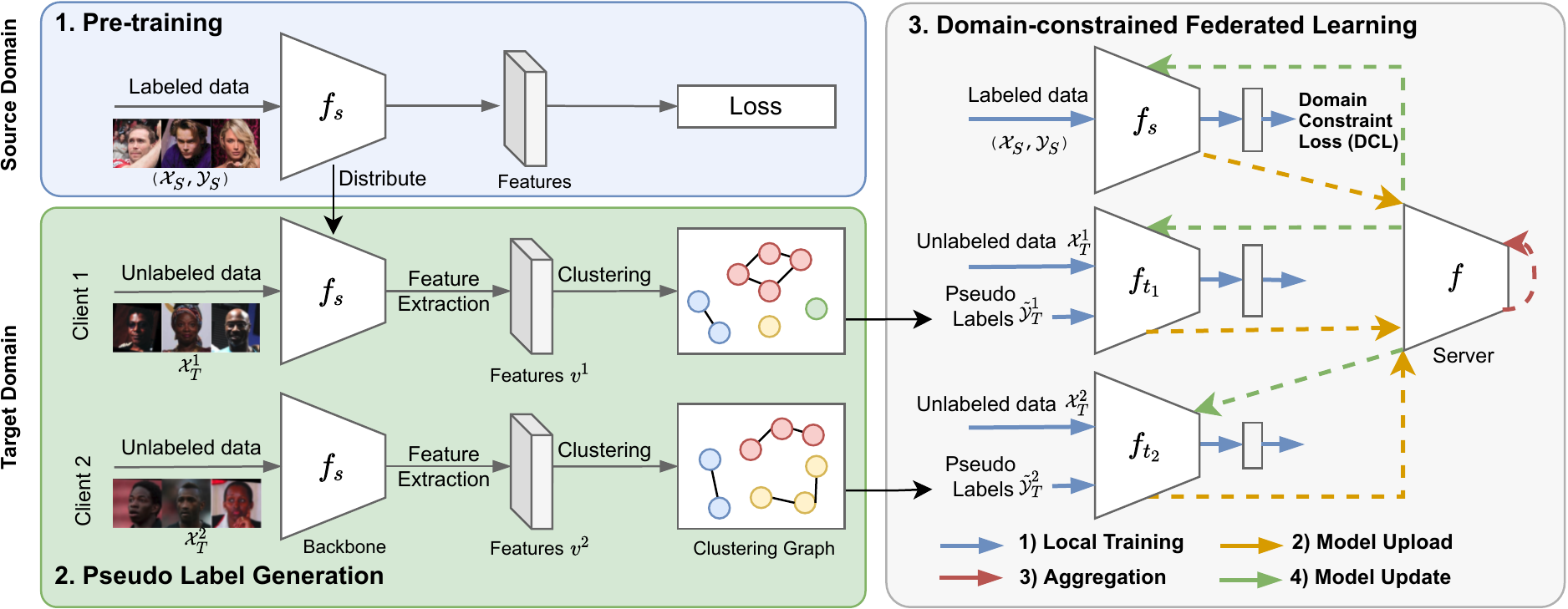}
\end{center}
   \caption{Overview of our proposed unsupervised federated face recognition approach (FedFR). FedFR addresses the domain shift problem under privacy constraints. We treat the source domain as one client and present two clients (e.g. edge devices) in the target domain for demonstration purposes. FedFR forms an end-to-end training pipeline with three stages: (1) \textit{Pre-training}: train a face recognition model $f_s$ in the source domain with data $\mathcal{D}_S=\{\mathcal{X}_S, \mathcal{Y}_S\}$; (2) \textit{Pseudo label generation}: predict pseudo labels $\mathcal{\tilde{Y}}_T$ for the unlabeled target domain data $\mathcal{D}_T=\{\mathcal{X}_T\}$ using clustering; (3) \textit{Domain-constrained Federated learning}: train a model $f$ that generalizes in the target domain collectively with clients in two domains. We propose a new domain constraint loss (DCL) to regularize the training in the source domain.}
\label{fig:framework}
\end{figure*}


\section{Related Work}
\subsection{Federated Learning}

Federated learning (FL) is a distributed training method that trains models with decentralized clients under the coordination of a server by transmitting model updates instead of raw data to preserve data privacy \cite{fedavg, kairouz2019fl-advances-open}. Previous studies have applied FL to computer vision tasks like image segmentation \cite{sheller2018brain-tumor2, li2019brain-tumor1}, object detection \cite{luo2019real}, and person re-identification \cite{zhuang2020fedreid}, but its application in face recognition is largely underexplored. The standard federated algorithm, FederatedAveraging (FedAvg) \cite{fedavg} requires identical models in the server and clients. While Federated Partial Averaging (FedPav) \cite{zhuang2020fedreid} is a federated algorithm for synchronizing only a fraction of the model between the server and the clients. We exploit the advantage of FedPav in this work by only synchronizing the backbone of face recognition models.


Recent works of FL mostly focus on supervised learning \cite{ caldas2018leaf, zhuang2020fedreid, fedma, fedprox}. Only a few of them study unsupervised federated learning: Peng et al. \cite{Peng2020ufda} proposed FADA for unsupervised federated domain adaptation; Song et al. \cite{song2020uda-federated-privacy} adopted maximum mean discrepancies (MMD) and homogeneous encryption for privacy-preserving unsupervised domain adaptation; van Berlo et al. \cite{van2020url-federated-edge} proposed software architecture for unsupervised representation learning. However, these methods may fail in face recognition because they are designed for tasks like classification. Unlike classification tasks that classes are identical between domains, face identities are different in two domains. We introduce an enhanced clustering algorithm to generate pseudo identities for unlabeled face images in the target domain.



\subsection{Unsupervised Domain Adaptation for Face Recognition} 

Unsupervised domain adaptation (UDA) for face recognition aims to deliver a high-performance face recognition model for the unlabeled target domain.
UDA has received great attention \cite{long2015dan, ganin2016dann, tzeng2017adda, long2018cada, hoffman2018cycada} recently. Existing research on UDA for face recognition mainly leverages some of these methods: Sohn et al. learned domain-invariant features through domain adversarial discriminator \cite{sohn2017uda-FR-video}. Luo et al. applied maximum mean discrepancies (MMD) loss to face recognition \cite{luo2018uda-FR}; Wang and Deng proposed a clustering-based method with MMD loss \cite{wang2020uda-clustering-face}. However, these approaches require locating data together, violating the constraint of data sharing between the source and target domains. In this work, we propose a new technique that leverages FL and enhances the FL algorithms to improve the performance in the target domain with privacy guaranteed.


\section{Methodology}

In this section, we introduce the proposed unsupervised federated face recognition approach (FedFR) to address the domain discrepancy between the source and target domains without data sharing.

\subsection{Problem Definition}

Before illustrating the details of FedFR, we present the problem and the assumptions first. The domain shift problem is introduced by training a model in one domain but deploying it to another. These two domains have different data distributions. This paper investigates the domain shift problem under the constraint that data is non-shareable between domains due to privacy protection and data is unbalanced. Specifically, we aim to obtain a model $f$ that delivers high performance in the target domain, given labeled source domain data $\mathcal{D}_S = \{\mathcal{X}_S, \mathcal{Y}_S\}$ and unlabeled target domain data $\mathcal{D}_T = \{\mathcal{X}_T\}$. $\mathcal{D}_S$ and $\mathcal{D}_T$ are under two assumptions: they are located in different places and not shareable; the size of $\mathcal{D}_T$ could be much smaller than $\mathcal{D}_S$. Moreover, in the target domain, data could be collected from cameras and stored in edge devices. Centralizing these face images to one server implies potential privacy leakage, so we assume that there are several non-shareable datasets in the target domain, $\mathcal{D}_T = \{\mathcal{D}_T^1, \cdots , \mathcal{D}_T^k, \cdots , \mathcal{D}_T^K | K \ge 1 \}$, where each client $k$ contains data $\mathcal{D}_T^k$. Thus, we use FedFR to train a face recognition model $f$ that generalizes well in the target domain, with $N=K+1$ decentralized clients, including one source domain client and $K$ target domain clients.

\subsection{Unsupervised Federated Face Recognition Overview}

Figure \ref{fig:framework} provides an overview of our proposed unsupervised federated face recognition approach (FedFR). To obtain a model that delivers high performance in the target domain without cross-domain data sharing, FedFR consists of three stages: source domain pre-training, pseudo label generation, and domain-constrained federated learning. These three stages form an end-to-end training pipeline: (1) We train a face recognition model $f_s$ with the source domain data $\{\mathcal{X}_S, \mathcal{Y}_S\}$. (2) Each client $k$ in the target domain downloads $f_s$, extracts features $v^k$ with local data $\mathcal{X}_T^k$, and generates pseudo labels $\mathcal{\tilde{Y}}_T^k$ by clustering $v^k$. (3) We conduct federated learning with a server to coordinate multiple clients --- the source domain as a client with $(\mathcal{X}_S, \mathcal{Y}_S)$ and all clients in the target domain with $(\mathcal{X}_T^k, \mathcal{\tilde{Y}}_T^k)$ --- to obtain a global model $f$ that generalizes across different domains. We summarize the FedFR in Algorithm \ref{algo:fedfr}.


\subsection{Pseudo Label Generation}
\label{sec:clustering}

To tackle the unlabeled data in the target domain, we generate pseudo labels for them using the source domain model and \textit{clustering algorithms}. As illustrated in the second stage in Figure \ref{fig:framework}, we first use the pre-trained model $f_s$ from the source domain to extract features $v^k$ from unlabeled data $\mathcal{D}_T^k$ in each client $k$, and then apply clustering algorithms on $v^k$ to form clustering graphs. \textit{Face images in the same cluster are regarded as having the same identity, tagged with the same pseudo label}.

The quality of pseudo labels depends on the performance of the clustering algorithm, so a good clustering algorithm is critical to FedFR. Built on FINCH \cite{sarfraz2019finch}, we propose an enhanced hierarchical clustering algorithm, Conditional FINCH (C-FINCH), by adding a new distance regularization when merging two clusters. Regarding each extracted feature as a cluster at the start, C-FINCH merges two clusters if they are first neighbors. Two clusters are first neighbors if distances of their centroids are the shortest or they share the same closest cluster. The distances measure the similarity between two clusters (identities) --- larger distance indicates that two identities are more divergent. Merging two clusters purely based on first neighbors would lead to unsatisfactory results because the distance of the first neighbors could be large. Hence, to optimize the clustering results, we tighten the condition of first neighbors by enforcing the distance of first neighbors to be \textit{smaller than a threshold $d$}. It ensures that only sufficiently similar features would be clustered together.

Compared with other popular clustering algorithms like K-means \cite{macqueen1967k-means} and DBSCAN \cite{ester1996dbscan}, C-FINCH does not require prior knowledge on the number of clusters, which would be hard to obtain in reality. Besides, unlike unsupervised learning studies that employ iterative training and clustering \cite{lin2019buc}, C-FINCH is much more efficient in computation as it only needs to predict pseudo labels once.

\SetKwInput{KwInput}{Input}                
\SetKwInput{KwOutput}{Output}              
\SetKwFunction{FnClient}{}
\SetKwFunction{FnServer}{}

\begin{algorithm}[t]
    \caption{Unsupervised Federated Face Recognition (FedFR)}
    \label{algo:fedfr}
    \SetAlgoLined
    \KwInput{Source domain data $\{\mathcal{X_S}, \mathcal{Y_S}\}$, target domain data $\mathcal{X}_T$, local iterations $E$, regularization intensity $\lambda$, training rounds $R$, total number of clients $N$}
    
    \KwOutput{Global model $f$ with parameters $\theta_R$}
    \textbf{Stage 1}: Pre-training \
    
    Train a model $f_s$ in the source domain with training data $\{\mathcal{X_S}, \mathcal{Y_S}\}$ and distribute it to the target domain\;
    
    \
    
    \textbf{Stage 2}: Pseudo label generation \
    
    \For{each client k in the target domain}{
        Extract features $v^k$ from data $\mathcal{X}_T$ using $f_s$ \;
        Obtain pseudo labels $\mathcal{\tilde{Y}}_T$ by clustering $v^k$ \;
    }
    
    \
    
    \textbf{Stage 3}: Federated Learning \
    
    $\theta_r$ is initialized with parameters $\theta^s$ of the model $f_s$ \;
    
    \For{each round r = 0 to R-1}{
        \textbf{Client:} \
        
        Each client $k$ initializes model with $\theta_r^k = \theta_r$ \;
        
        \For{each iteration e = 0 to E-1}{
            Source domain client trains on $\theta_r^k$ with the domain constrained loss (DCL) in Equation \ref{eq:regularization-l2} \;
            Target domain clients train on $\theta_r^k$\;
        }
        Each client obtains model $\theta_{r+1}^k$ and sends it to the server\;
        
        \textbf{Server:} \
        
        Server aggregates client models with $\theta_{r+1} = \frac{1}{N} \sum_{k=1}^N \theta_{r+1}^k$ \;
    }    

\end{algorithm}

\subsection{Domain-constrained Federated Learning}
\label{sec:federated-learning}

To reduce the domain discrepancy without sharing data between domains, we transfer the knowledge from the source domain to the target domain via federated learning (FL). Regarding the source domain and each edge of the target domain as clients, they perform training collaboratively with the coordination of a central server. As the source domain normally contains much more data than the target domain, we propose to regularize the source domain training with a new \textit{domain constraint loss} (DCL). The whole training process is termed \textit{domain-constrained federated learning}. 

The third stage in Figure \ref{fig:framework} presents the training flow of domain-constrained FL. Training begins with initializing all clients in both domains with model parameters $\theta^s$ of the pre-trained model $f_s$ from the source domain. It conducts iterative training, where each training round $t$ includes the following steps: (1) \textit{Local Training}: each client conducts $E$ local iterations of training with its local dataset. The source domain client trains with $(\mathcal{X}_S, \mathcal{Y}_S)$ using DCL, while each target domain client $k$ trains with $(\mathcal{X}_T^k, \mathcal{\tilde{Y}}_T^k)$. (2) \textit{Model Upload}: each client sends the model training updates $\theta_{r+1}^k$ to the central server. (3) \textit{Aggregation}: the server aggregates the model updates to obtain a new global model with $\theta_{r+1} = \frac{1}{N} \sum_{k=1}^N \theta_{r+1}^k$ . (4) \textit{Model Update}: each client downloads the global model $\theta_{r+1}$ to update its local model for the next round of training.

\begin{figure}[t]
\begin{center}
\includegraphics[width=0.9\linewidth]{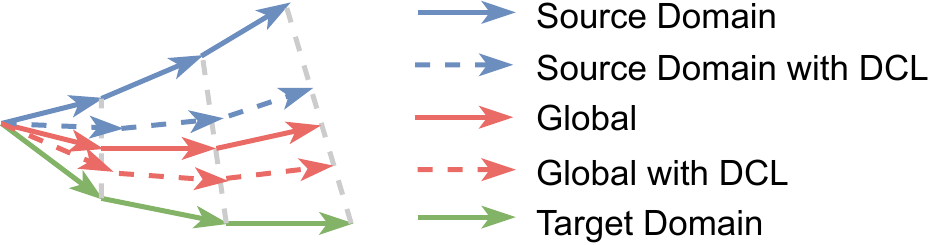}
\end{center}
   \caption{Illustration of domain constrained loss (DCL). DCL addresses unbalanced data volume between domains by regularizing the source domain such that the global model can lean towards the target domain.}
\label{fig:intuition}
\end{figure}

\begin{table*}[t]
    \caption{Statistics of two newly constructed benchmarks: FedFR-Small (FedFR-S) and FedFR-Large (FedFR-L). The benchmarks contain training data in (a) and evaluation data in (b). We simulate real-world scenarios using Caucasian identities for the source domain and African identities for the target domain. For evaluation, we use the Test set to evaluate face verification and construct queries and galleries to evaluate face identification.}
    \begin{subtable}[h]{0.38\textwidth}
        \setlength{\tabcolsep}{0.4em}
        \begin{center}
        \begin{tabular}{ccccccc}
        \toprule
        \multicolumn{1}{c}{\multirow{2}{*}{Benchmark}} &
        \multicolumn{2}{c}{Source Domain} &
        \multicolumn{1}{c}{} &
        \multicolumn{2}{c}{Target Domain}
        \\ 
        \cline{2-3} \cline{5-6}
        \multicolumn{1}{c}{} & \# IDs & \# Images & & \# IDs & \# Images 
        \\
        \midrule 
        
        FedFR-S & 7,000 & 326,484 & & 7,000 & 324,376 \\ 
        FedFR-L & 87,072 & 4,434,177 & & 7,000 & 324,376 \\ 
        
        \bottomrule
        \end{tabular}
        \end{center}
        \caption{Training data of the benchmarks}
        \label{tab:training-data}
    \end{subtable}
    \hfill
    \begin{subtable}[h]{0.6\textwidth}
        \setlength{\tabcolsep}{0.4em}
        \begin{center}
        \begin{tabular}{llccccccccc}
        \toprule
        \multicolumn{1}{l}{\multirow{2}{*}{Dataset}} &
        \multicolumn{1}{l}{\multirow{2}{*}{Domain}} &
        \multicolumn{1}{c}{} &
        \multicolumn{2}{c}{Test} &
        \multicolumn{1}{c}{} &
        \multicolumn{2}{c}{Query} &
        \multicolumn{1}{c}{} &
        \multicolumn{2}{c}{Gallery}
        \\ 
        \cline{4-5} \cline{7-8} \cline{10-11}
        & & & \# IDs & \# Images & & \# IDs & \# Images & & \# IDs & \# Images \\ 
        \midrule
        Caucasian & Source & & 2,959 & 10,196 & & 2,793 & 2,793 & & 2,958 & 6,387 \\ 
        African & Target & & 2,995 & 10,415 & & 2,865 & 2,865 & & 2,995 & 6,770 \\ 
        \bottomrule
        \end{tabular}
        \end{center}
        \caption{Evaluation data}
        \label{tab:evaluation-data}
    \end{subtable}
     \label{tab:temps}
\end{table*}

However, conventional FL (FedAvg) \cite{fedavg} is not directly applicable to our setting. We propose three enhancements as follows:

\textbf{Regularizing Source Domain} To tackle the unbalanced data volume of two domains, we propose DCL to regularize the source domain training. Before introducing the details of DCL, we first analyze the limitations of conventional FL, FedAvg \cite{fedavg}.


The objective of FedAvg is to obtain a global model $f$ with: 

\begin{equation}
    \min_{\theta \in \mathbb{R}^d} f(\theta) :=  \frac{1}{N} \sum_{k=1}^N F_k(\theta),
\label{eq:fl}
\end{equation}

where $N$ is the number of clients, and for client $k$, $F_k: \mathbb{R}^d \rightarrow \mathbb{R}$ represents the expected loss over data distribution $\mathcal{D}_k$:



\begin{equation}
    F_k(\theta) := \mathbb{E}_{x_k \sim \mathcal{D}_k}[f_k(\theta;x_k)],
\end{equation}

where $x_k$ is the data in client $k$ and $f_k(\theta;x_k)$ is a loss function to train model $\theta$.


FedAvg (Equation \ref{eq:fl}) aggregates model updates with averaging, which weighs the importance of the source and target domains equally. 
However, it is not optimal to deliver high performance in the target domain, especially when the target domain contains \textit{much less} data than the source domain. As a result, we need to regularize the source domain and reinforce the importance of the target domain.
Inspired by FedProx \cite{fedprox}, which adds a proximal term to all clients, we propose DCL by adding a regularized $l_2$-norm in the source domain. The new local objective of the source domain training is:


\begin{equation}
     \min_{\theta \in \mathbb{R}^d} h(\theta; \theta_r) := F(\theta) + \frac{\lambda}{2} \left\Vert \theta - \theta_r \right\Vert^2,
\label{eq:regularization-l2}
\end{equation}


where $\theta$ is the source domain model to be optimized, $\theta_r$ is the parameters of the global model from the previous round, and $\lambda$ is a control parameter for the intensity of the regularization. The local objective of the target domain clients remains unchanged. 

We depict the intuition of DCL in Figure \ref{fig:intuition}. In each training round, the global model is obtained by averaging the source and target models. We add DCL in the source domain to regularize its training so that it does not deviate far from the global model. This is especially relevant when the source domain contains much more data. With DCL in the source domain, the global model can lean towards the target domain model, resulting in better performance on the target domain. 





\textbf{Local Iterations} Besides DCL, we enhance FedAvg \cite{fedavg} by using \textit{local iteration} as the minimum execution unit instead of local epoch. FedAvg is a synchronous protocol, where the server waits for all clients to train $L$ local epochs before proceeding to the next round. $L$ epochs mean passing through the training data for $L$ times. However, as the source domain could contain much more data, the server and the target domain clients need to wait for the source domain client to finish training, which imposes long training time and wastes computation power. Instead of training $L$ local epochs, we propose to train $E$ local iterations in clients in each training round, such that clients in both domains can execute the same amount of computation and finish training simultaneously. We consider the minimum $E$ to be 1.5k, which is close to one epoch in the target domain clients. The impact of different values of $E$ is analyzed in Section \ref{sec:analysis-of-fedfr}. When the source domain contains $J$ times more data than the target domain, using local iterations only needs $\frac{1}{J}$ of the original training time, assuming that the training time of one batch is the same in all clients. We provide more details about reductions in training time in the supplementary.



\textbf{Federated Partial Averaging} We also enhance FedAvg to only synchronize partial model in each client. A face recognition model comprises a backbone and a loss. The model structures of the loss depend on the number of identities of datasets, so they are not identical in all clients. However, FedAvg requires identical models, which is incompatible with face recognition. Inspired by Federated Partial Averaging (FedPav) \cite{zhuang2020fedreid} that synchronizes models without classifiers for person re-identification, we enhance FedAvg to only synchronize the backbone --- without the neural network of the loss. It is also reasonable because the backbone contains low-level representations of face images while the neural network of the loss contains high-level features that vary according to datasets. The global model $f$, mentioned in previous sections, stands for the backbone of a model.

\begin{table}[t]
\caption{Face verification (ver.) accuracy and identification (id.) at rank-1 on the \textit{FedFR-S} benchmark. Although FedFR does not outperform \textit{supervised training} containing target dataset (Target-Only and Merge), it effectively improves the performance on the target domain (African) and outperforms other methods, with setting $K=4$ and $\lambda=0.002$.}
\begin{center}
\begin{tabular}{lcccccc}
\toprule
\multicolumn{1}{l}{\multirow{2}{*}{Model}} &
\multicolumn{2}{c}{African (\%)} &
\multicolumn{1}{c}{} &
\multicolumn{2}{c}{Caucasian (\%)} \\ 
\cline{2-3} \cline{5-6}
\multicolumn{1}{c}{} & Ver. & Id. & & Ver. & Id.\\ 
\midrule
Source-Only & 70.40 & 26.42 & & 89.10 & 71.79 \\ 
Target-Only & 83.35 & 73.54 & & 86.23 & 46.26 \\ 
Merge & \textbf{89.40} & \textbf{84.75} & & \textbf{92.75} & \textbf{85.43} \\ 
Fine-tune & 81.07 & 64.08 & & 85.42 & 67.60 \\ 
DAN & 73.57 & 31.62 & & 85.53 & 62.37 \\ 
DANN & 72.38 & 31.06 & & 89.15 & 70.49 \\ 
\hline
FedFR (Ours) & 82.50 & 70.47 & & 87.62 & 77.01 \\  

\bottomrule
\end{tabular}
\end{center}
\label{tab:fedfr-s}
\end{table}

\begin{table*}[t]
\caption{Performance of face verification and identification (Id.) on the \textit{FedFR-L} benchmark. For verification, we present accuracy and true acceptance rate (TAR) at different false acceptance rates (FAR) of 0.1, 0.01, and 0.001. FedFR significantly improves the performance on the target domain (African), even outperforming the Merge model. It almost maintains the performance on the source domain (Caucasian) at the same time, with setting $K=1$ and $\lambda=0.01$.}
\begin{center}
\begin{tabular}{lcccccccccccccc}
\toprule
\multicolumn{1}{l}{\multirow{3}{*}{Model}} &
\multicolumn{6}{c}{African (\%)} &
\multicolumn{1}{c}{} &
\multicolumn{6}{c}{Caucasian (\%)} 
\\ 
 \cline{2-7}  \cline{9-14}
\multicolumn{1}{c}{} &

\multicolumn{4}{c}{Verification} &
\multicolumn{1}{c}{} &
\multicolumn{1}{c}{Id.} &

\multicolumn{1}{c}{} &

\multicolumn{4}{c}{Verification} &
\multicolumn{1}{c}{} &
\multicolumn{1}{c}{Id.}
\\ 
\cline{2-5} \cline{7-7} \cline{9-12} \cline{14-14} 
\multicolumn{1}{c}{} & Acc & 0.1 & 0.01 & 0.001 & & Rank-1 & & Acc & 0.1 & 0.01 & 0.001 & & Rank-1 \\
\midrule
Source-Only & 86.65 & 83.27 & 55.93 & 37.80 & & 82.13 & & 94.47 & 95.67 & \textbf{87.53} & \textbf{76.67} & & 93.52 \\ 
Target-Only & 83.35 & 75.6 & 41.80 & 22.00 & & 73.54 & & 86.23 & 81.93 & 54.83 & 38.27 & & 46.26 \\
Merge & 88.60 & 87.00 & 66.90 & 48.10 & & 89.04 & & \textbf{94.78} & \textbf{96.60} & 85.63 & 74.43 & & 94.38 \\ 
Fine-tune & 88.72 & 87.10 & 62.23 & 35.95 & & 84.05 & & 87.73 & 85.87 & 62.87 & 39.77 & & 77.77 \\ 
DAN & 86.98 & 83.87 & 61.73 & 41.43 & & 83.18 & & 94.07 & 95.43 & 85.20 & 72.13 & & 93.30 \\ 
DANN & 86.77 & 83.23 & 60.80 & 44.90 & & 82.23 & & 93.98 & 96.03 & 86.23 & 73.97 & & 93.59  \\ 
\hline

FedFR (Ours) & \textbf{91.55} & \textbf{91.57} & \textbf{76.97} & \textbf{49.40} & & \textbf{93.26} & & 94.25 & 96.00 & 85.60 & 70.93 & &  \textbf{94.99} \\ 

\bottomrule
\end{tabular}
\end{center}
\label{tab:fedfr-l}
\end{table*}

\section{Experiments}
\label{sec:experiments}

In this section, we first define two benchmarks for FedFR. Then we present the benchmark results, analysis of FedFR, and ablation studies on these two benchmarks.



\subsection{FedFR Benchmark}
\label{sec:benchmark}

We construct two benchmarks for federated face recognition since little work has been devoted to it, except mentioned in \cite{hu2020oarf}. We design the source and target domains to contain datasets with different races --- \textit{Caucasian in the source domain and African in the target domain}, simulating models trained in one region and deployed into another. To better evaluate the effectiveness of algorithms, these two benchmarks vary the amount of training data in the source domain, representing different capabilities of the pre-trained models. The details of the datasets are shown in Table ~\ref{tab:training-data}. These benchmarks also include evaluation datasets for each domain in Table \ref{tab:evaluation-data}.


\textbf{FedFR-S} FedFR-Small benchmark (FedFR-S) contains 7K labeled Caucasian identities in the source domain and 7K unlabeled African identities in the target domain, adopting from race-balanced dataset BUPT-Balancedface \cite{wang2020bupt}. 

\textbf{FedFR-L} FedFR-Large benchmark (FedFR-L) contains around 87K labeled identities in the source domain and 7K unlabeled African identities in the target domain. In comparison with FedFR-S, FedFR-L is more representative of the real-world scenario that a face recognition model, trained with a large amount of the source domain data, performs well in the source domain but degrades in the target domain.
The source domain data of FedFR-L is built from MS-Celeb-1M (MS1M) \cite{guo2016ms1m}. MS1M is a large-scale dataset, containing 100K identities, 82\% of which are Caucasian. 
We construct FedFR-L source domain data by excluding related data that are originally from MS1M: 7K African identities of the target domain and around 6K identities of the evaluation dataset. Although FedFR-L still contains around 6.5\% of East Asian identities according to \cite{sohn2018uda-metric-learning}, it is similar to the real-world scenario that the training data would not purely belong to the same domain for better generalization.

\textbf{Evaluation} We evaluate the verification and identification performance of face recognition models using data from Racial Faces in-the-Wild (RFW) dataset \cite{wang2019rfw}, as shown in Table \ref{tab:evaluation-data}. We use around 3K Caucasian identities and around 3K African identities to evaluate the performance of the source and target domains, respectively. For face verification, we present the verification accuracy and true acceptance rate (TAR) at false acceptance rates (FAR) of 0.1, 0.01, and 0.001. For face identification, we report the rank-1 accuracy by matching a query to a gallery of images. We use the whole test set to evaluate face verification performance and construct queries and galleries to evaluate face identification performance. The details of the construction are described in the supplementary.

\subsection{Implementation Details}

We implement FedFR and conduct the experiments using PyTorch \cite{paszke2019pytorch}. For image preprocessing, we transform the face images with augmentations, and then crop and resize the face images to the size of 112 x 112. For network architecture, we choose ResNet-34 \cite{he2016resnet} as the backbone and the ArcFace \cite{deng2019arcface} as the loss. The dimension of features extracted from the backbone is 256. All experiments are conducted in clusters of eight NVIDIA V100 GPUs.  

In FedFR, we simulate FL by training each client in a GPU and use the PyTorch GPU communication backend to simulate the server aggregation. FedFR includes one source domain client by default. For the target domain, we split the 7K African identities evenly to $K$ partitions to simulate $K$ clients. For example, when simulating 4 clients in the target domain, each client would contain 1,750 unlabeled data of African identities. Besides, we tune the best regularization intensity $\lambda$ for different values of $K$ in both benchmarks. By default, we set $K = 4$ and $\lambda = 0.002$ for the FedFR-S benchmark, and $K = 1$ and $\lambda = 0.01$ for the FedFR-L benchmark. We further analyze the impact of $K$ in Section \ref{sec:analysis-of-fedfr}. For clustering, we set the distance regularization of C-FINCH $d = 1.2$ for the FedFR-S benchmark and $d = 0.9$ for the FedFR-L benchmark.





\subsection{Benchmark Results}

We compare FedFR with models obtained by other methods on the FedFR-S and FedFR-L benchmarks. The following are the compared models: 
(1) \textit{Source-Only}: the baseline model obtained from supervised training with the source domain data. 
(2) \textit{Target-Only}: the model obtained from supervised training with the target domain data.
(3) \textit{Merge}: the model obtained from supervised training with data from both the source and target domains, presenting possible upper bound without data sharing constraint.
(4) \textit{Fine-tune}: the model fine-tuned from the baseline model with pseudo labels generated from the clustering algorithm. 
(5) \textit{DAN} \cite{long2015dan}: the model trained by domain adaptation network with maximum mean discrepancies (MMD) loss, which is the classic discrepancy-based domain adaptation method.
(6) \textit{DANN} \cite{ganin2016dann}: the model trained by domain adversarial neural network, which is the classic adversarial-based domain adaptation method.
(7) \textit{FedFR}: the model trained using all components in FedFR.



\textbf{FedFR-S Comparisons} Table \ref{tab:fedfr-s} compares models trained using different methods on the FedFR-S benchmark. The performance discrepancy between African and Caucasian of the \textit{Source-Only} model is $\sim$19\% while the difference in the \textit{Merge} model is no more than 3\%, \textit{revealing the severe impact of the domain shift problem}. FedFR tackles the problem, improving $\sim$12\% on verification accuracy and $\sim$44\% on rank-1 accuracy of the target domain (African).
Although it does not outperform the upper bound (\textit{Merge}), it is close to the \textit{Target-Only} model and is superior to \textit{Fine-tune}, DAN, and DANN models. Additionally, fine-tuning boosts the performance on African dataset, but it suffers from catastrophic forgetting \cite{kirkpatrick2017catastrophic-forgetting} as shown in Figure \ref{fig:source-domain}. With the progress of more iterations of training on the African dataset, the model gradually forgets the knowledge learned from the Caucasian dataset,
causing performance decreases on Caucasian evaluation. We run the experiments with $K = 4$ and $\lambda = 0.002$ for DCL. We provide more results of verification rate at FAR=0.1, 0.01, and 0.001 in the supplementary. 



\textbf{FedFR-L Comparisons} Table \ref{tab:fedfr-l} reports the experiment results on the FedFR-L benchmark with verification accuracy, verification rate at FAR=0.1, 0.01, and 0.001, and identification at rank-1. The impact of the domain shift problem is still significant, which leads to a $\sim$6\% performance gap on verification accuracy between these two domains in the baseline. What stands out in the table is the performance of FedFR, which outperforms all other models on the target domain (African), at the same time, maintains good performance on the source domain. It demonstrates the significance of our method. On the contrary, despite that DAN and DANN retain the performance on the Caucasian dataset, they hardly improve the performance on the African dataset.

\subsection{Analysis of FedFR}
\label{sec:analysis-of-fedfr}

This section first studies the communication cost and scalability of FedFR. We then analyze the impact of the clustering algorithm and source domain on the performance of FedFR. 

\begin{figure}
    \begin{subfigure}{.48\linewidth}
        \centering
        \includegraphics[width=\linewidth]{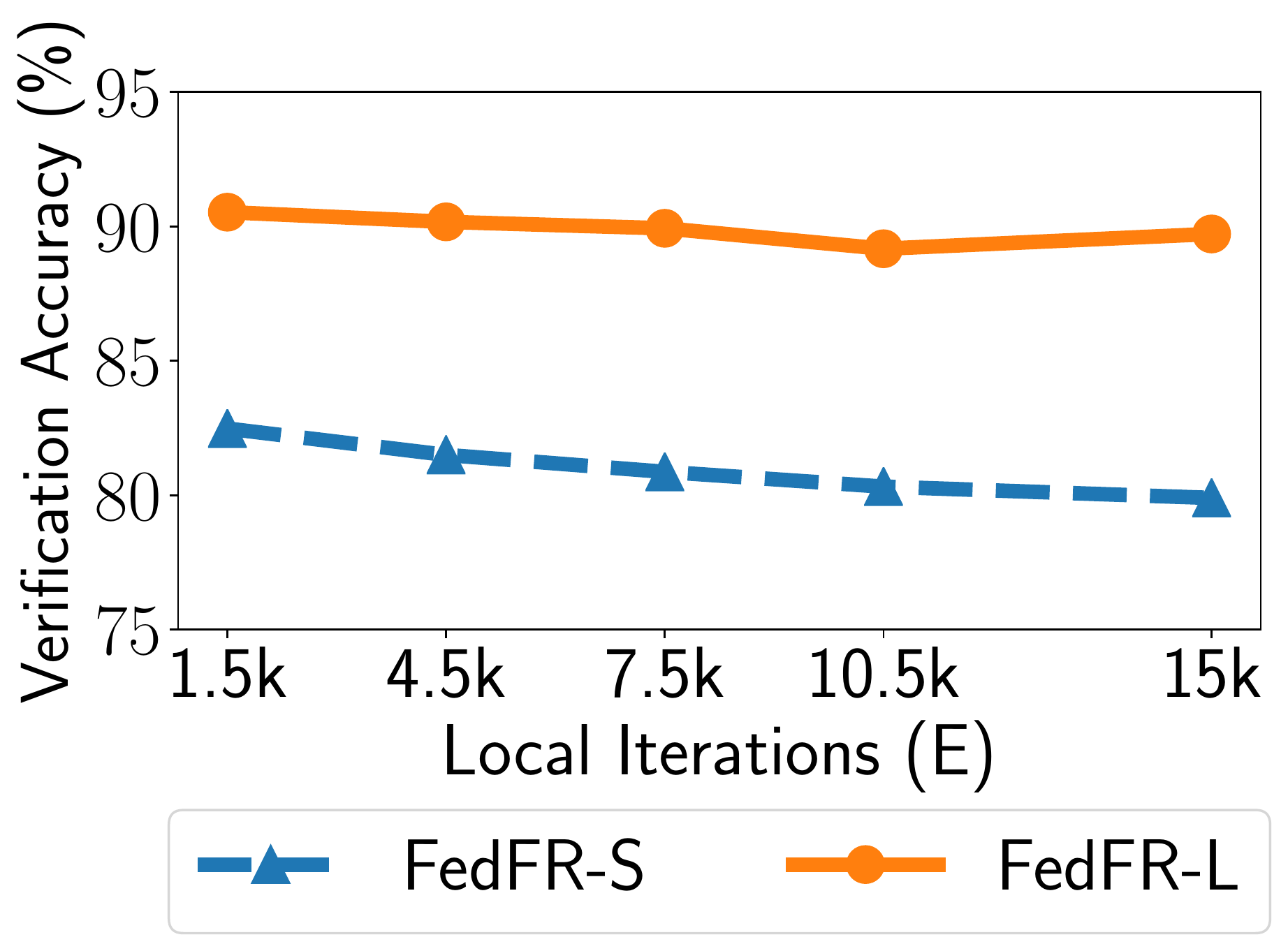}
        \caption{Verification Accuracy}
    \label{fig:local-iter-acc}
    \end{subfigure}%
    \hfill
    \begin{subfigure}{.48\linewidth}
        \centering
        \includegraphics[width=\linewidth]{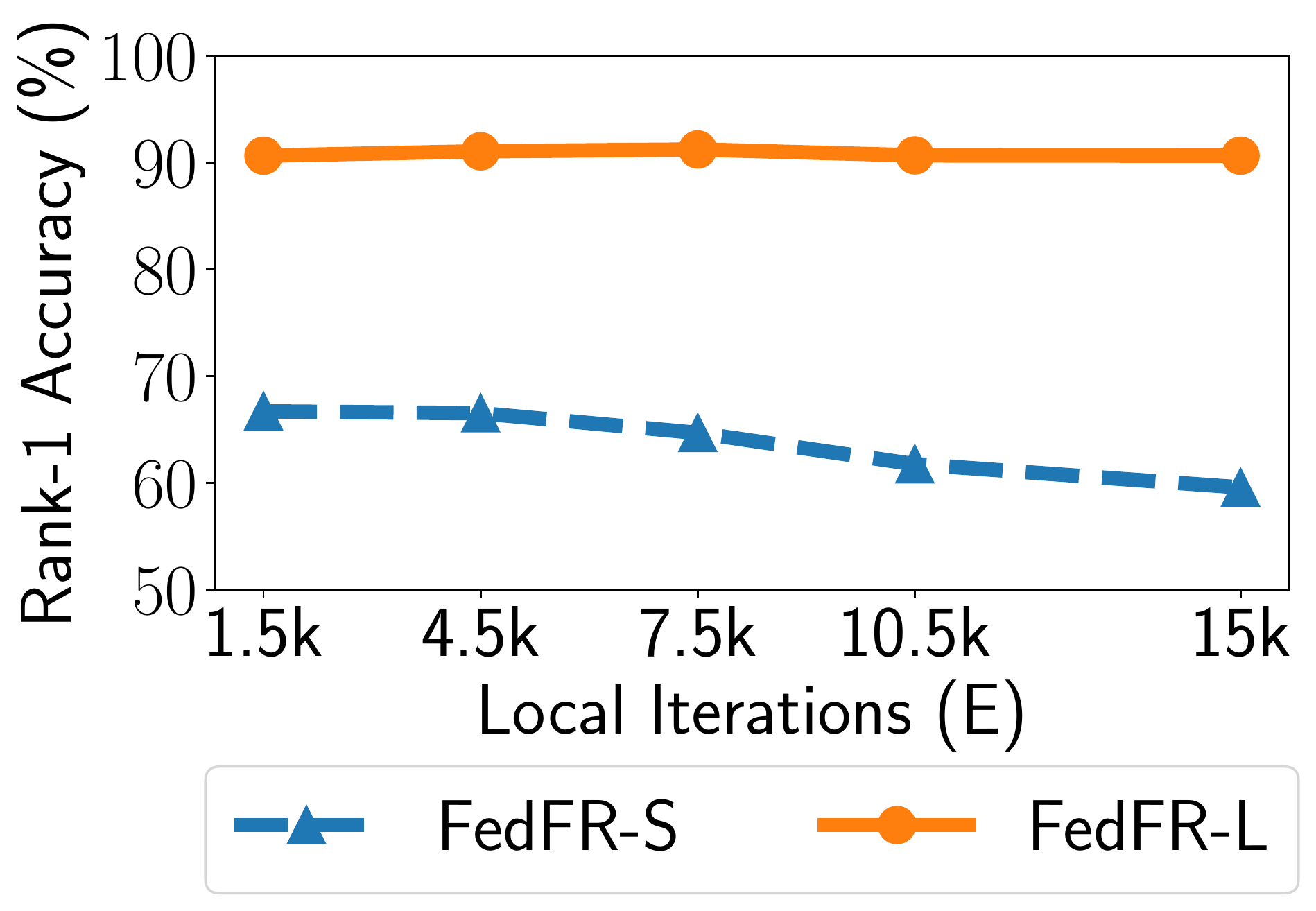}
        \caption{Identification at Rank-1}
        \label{fig:local-iter-rank}
    \end{subfigure}

    \caption{Impact of local iterations $E$ on (a) verification and (b) identification accuracy in both FedFR-S and FedFR-L benchmarks with $K = 4$. Increasing local iterations reduces performance in the FedFR-S benchmark, but the performance in the FedFR-L benchmark is only slightly affected.}
    \label{fig:local-iterations}
\end{figure}

\textbf{Communication Cost} The number of local iterations $E$ determines the communication cost of FedFR. As the data size transmitted between server and clients is the same in every communication, the communication cost depends on the communication frequency. With total $M$ number of training iterations, the communication frequency $F = \frac{M}{E}$. A larger value implies lower communication costs. Figure \ref{fig:local-iterations} compares performances of various values of $E$ on the FedFR-S and FedFR-L benchmarks, evaluating on the African dataset. On the FedFR-S benchmark, increasing the number of local iterations $E$ reduces performance, indicating the \textit{trade-off} between communication cost and performance when the data volume of the source domain is small. In contrast, on the FedFR-L benchmark, a larger $E$ causes only slight decreases in performance. It demonstrates that FedFR can maintain the performance with \textit{lower communication costs} by using larger $E$ in the real-world scenario where the source domain contains much more data. We train for total $M = 80k$ iterations in these experiments.

\begin{figure}
    \begin{subfigure}{.48\linewidth}
        \centering
        \includegraphics[width=\linewidth]{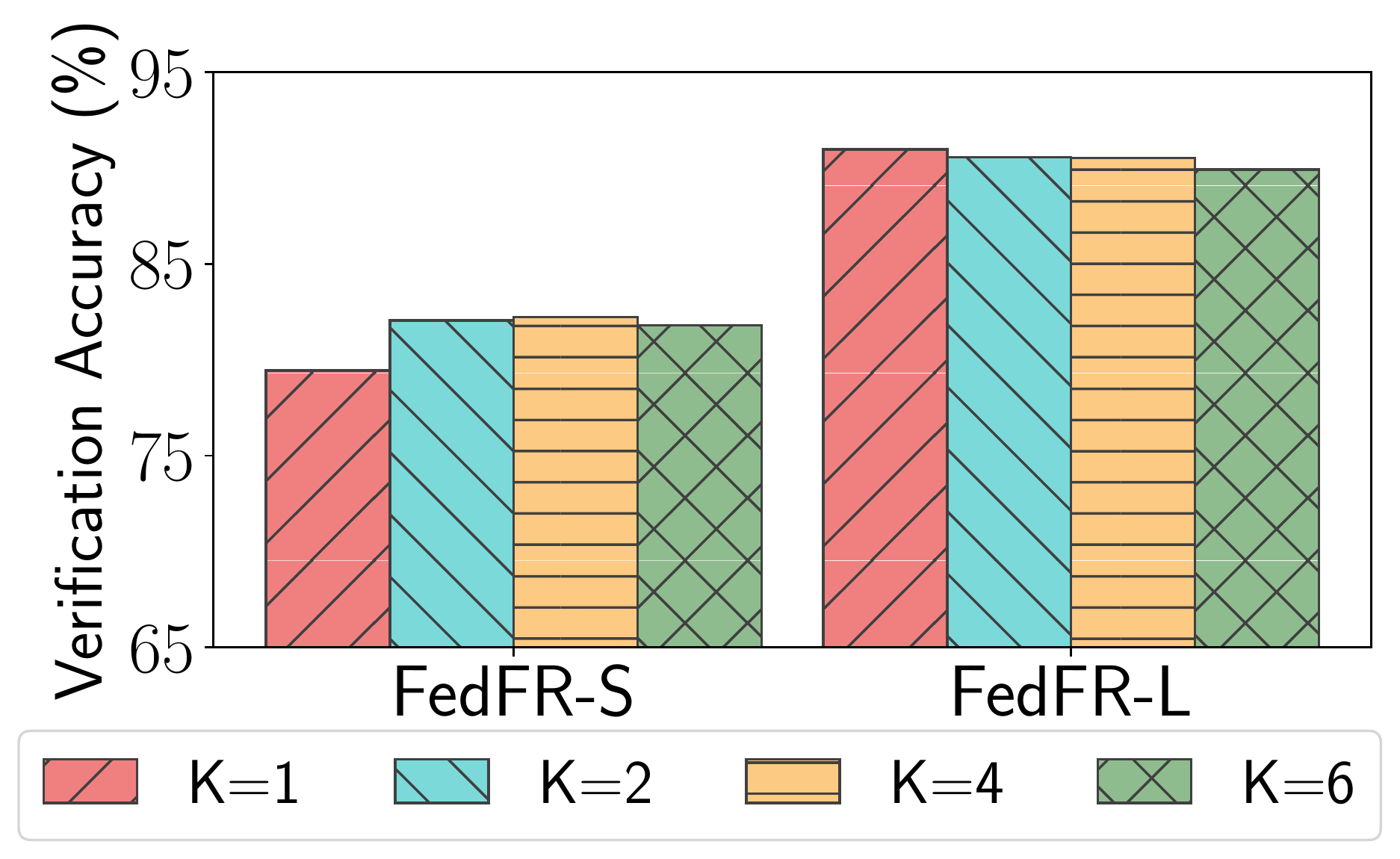}
        \caption{Verification Accuracy}
    \label{fig:number-of-acc}
    \end{subfigure}%
    \hfill
    \begin{subfigure}{.48\linewidth}
        \centering
        \includegraphics[width=\linewidth]{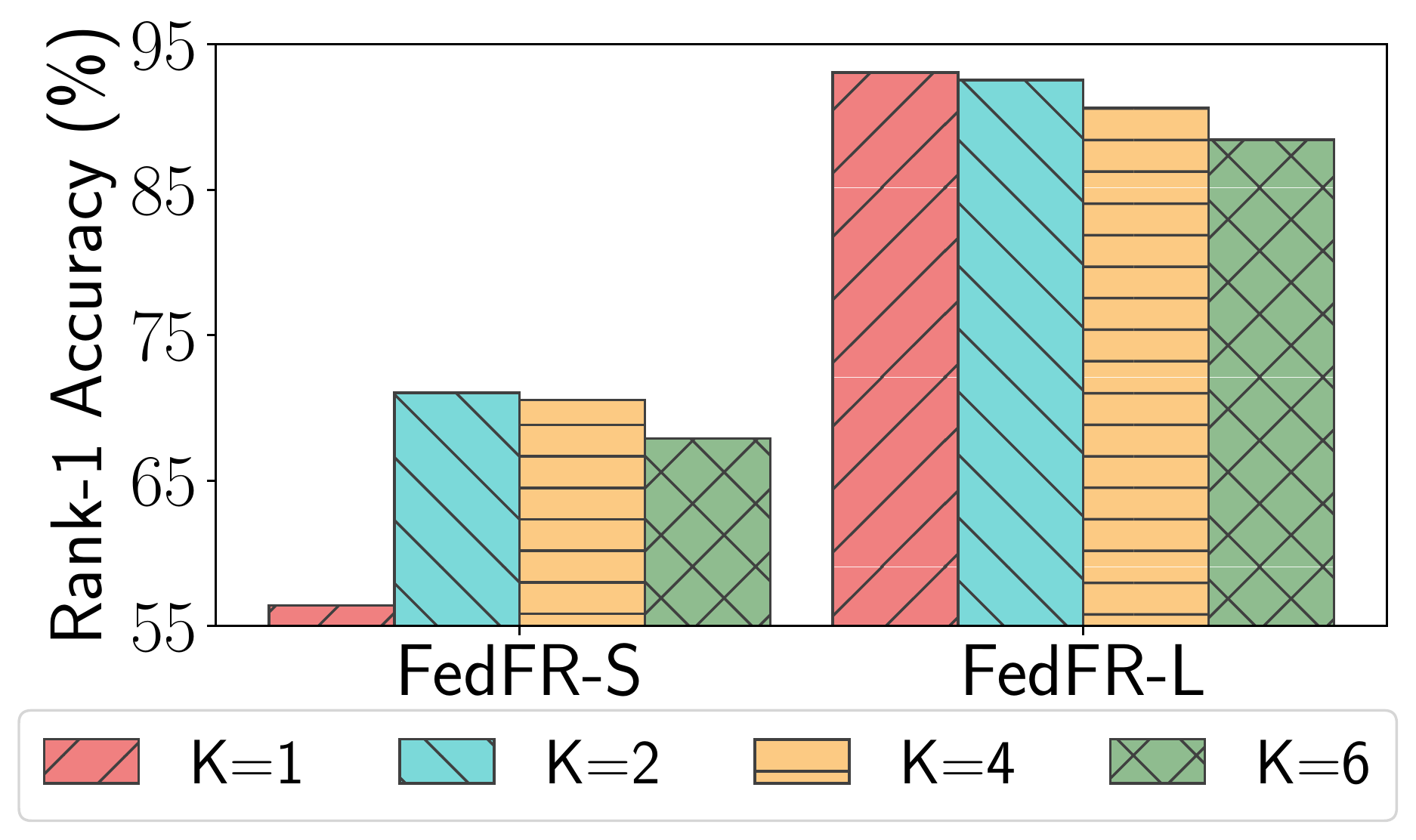}
        \caption{Identification at Rank-1}
        \label{fig:number-of-rank}
    \end{subfigure}

    \caption{Performance comparison of different numbers of target domain clients $K$ on the FedFR-S and FedFR-L benchmarks. A larger $K$ slightly affects the performance on the FedFR-L benchmark. While on the FedFR-S benchmark, $K = 6$ results in better performance than $K = 1$.}
    \label{fig:number-of-clients}
\end{figure}

\textbf{Scalability} FedFR is scalable to handle a larger number of target domain clients $K$. Figure \ref{fig:number-of-clients} compares the performance of $K = \{1, 2, 4, 6\}$. On the FedFR-L benchmark, although larger K slightly degrades the performance, the performance of $K = 6$ is still better than other methods compared in the paper. In contrast, on the FedFR-S benchmark, the performance of larger $K$ is better $K = 1$. All these results demonstrate that FedFR supports data scattered in multiple edge devices instead of centralizing them. Due to computation constraints, we only evaluate $K$ up to 6. For scenarios with huge numbers of edge devices in the target domain, since edges share similar domains, we can explore two strategies: (1) Train a model with several edges using FedFR and deploy it to the rest; (2) Split edges into smaller groups, and use FedFR to conduct training among each group.

\begingroup
\setlength{\tabcolsep}{0.55em}
\begin{table}[t]
\caption{Performance comparison of different clustering algorithms. Our enhanced C-FINCH outperforms K-means and DBSCAN even without relying on prior knowledge of the dataset. It greatly outperforms FINCH on the FedFR-L benchmark.}
\begin{center}
\begin{tabular}{clccccccc}
\toprule
\multicolumn{1}{c}{\multirow{2}{*}{Benchmark}} &
\multicolumn{1}{l}{\multirow{2}{*}{Algorithm}} &
\multicolumn{1}{c}{\multirow{2}{*}{F-score (\%)}} &
\multicolumn{2}{c}{African (\%)}
\\ 
\cline{4-5} 
\multicolumn{1}{c}{} &
\multicolumn{1}{c}{} &
\multicolumn{1}{c}{} &
\multicolumn{1}{c}{Ver.} &
\multicolumn{1}{c}{Id.}
\\
\midrule
\multirow{4}{*}{FedFR-S} &
DBSCAN \cite{ester1996dbscan} & 51.04 & 61.18 & 15.43 \\
& K-means \cite{macqueen1967k-means} & 60.87 & 80.45 & 64.08 \\ 
& FINCH \cite{sarfraz2019finch} & 64.62 & 81.42 & 60.31 \\ 
& C-FINCH & \textbf{64.65} & \textbf{82.18} & \textbf{69.42} \\ 

\midrule
\multirow{4}{*}{FedFR-L} &
DBSCAN \cite{ester1996dbscan} & 84.42 & 86.75 & 88.73 \\
& K-means \cite{macqueen1967k-means} & 87.17 & 89.99 & 87.96 \\ 
& FINCH \cite{sarfraz2019finch} & 70.08 & 87.58 & 81.61 \\ 
& C-FINCH & \textbf{91.54} & \textbf{90.43} & \textbf{91.24} \\
\bottomrule
\end{tabular}
\end{center}
\label{tab:clusterings}
\end{table}
\endgroup

\begin{figure}
    \begin{subfigure}{.48\linewidth}
        \centering
        \includegraphics[width=\linewidth]{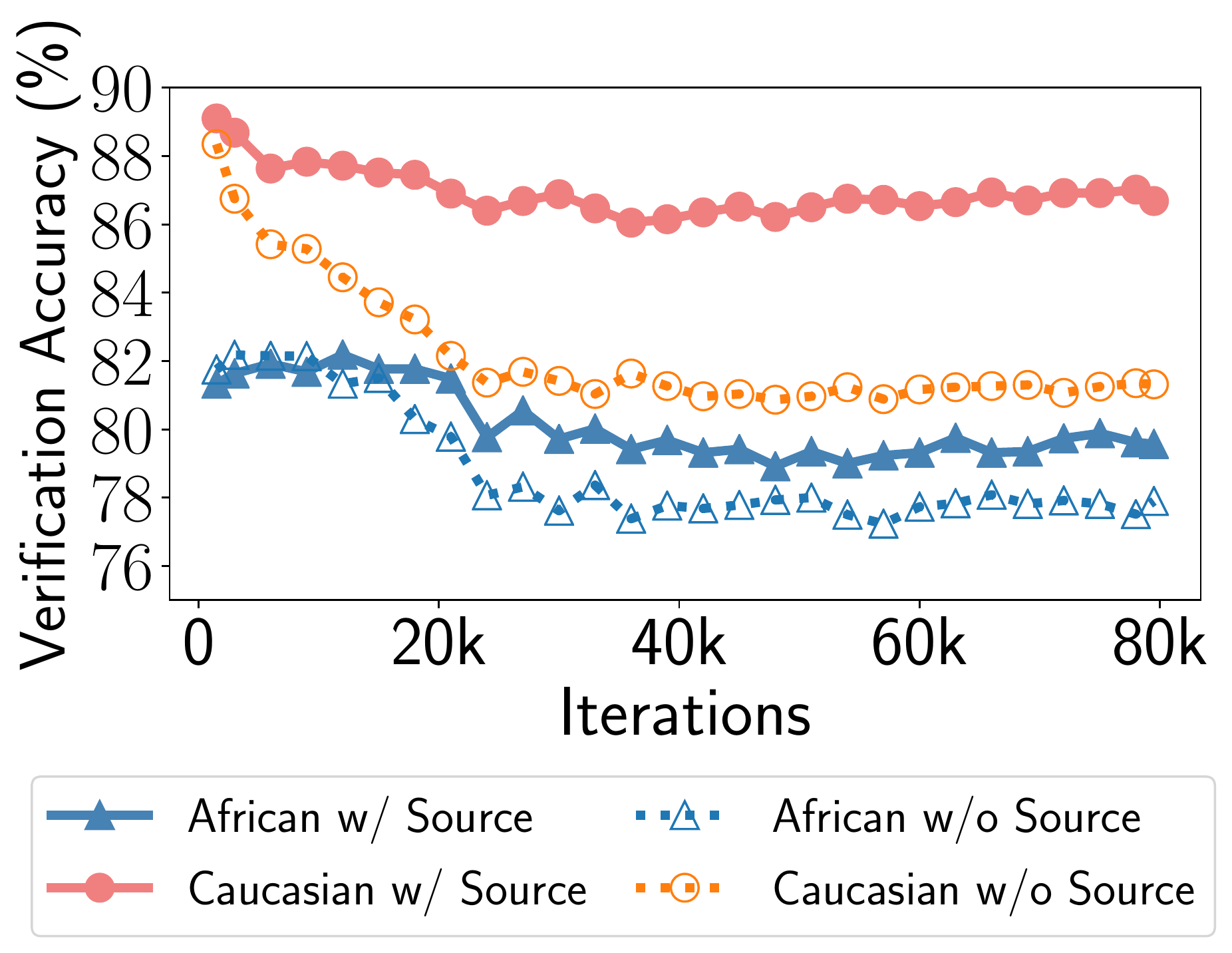}
        \caption{FedFR-S Benchmark}
        \label{fig:source-fedfr-s}
    \end{subfigure}
    \hfill
    \begin{subfigure}{.48\linewidth}
        \centering
        \includegraphics[width=\linewidth]{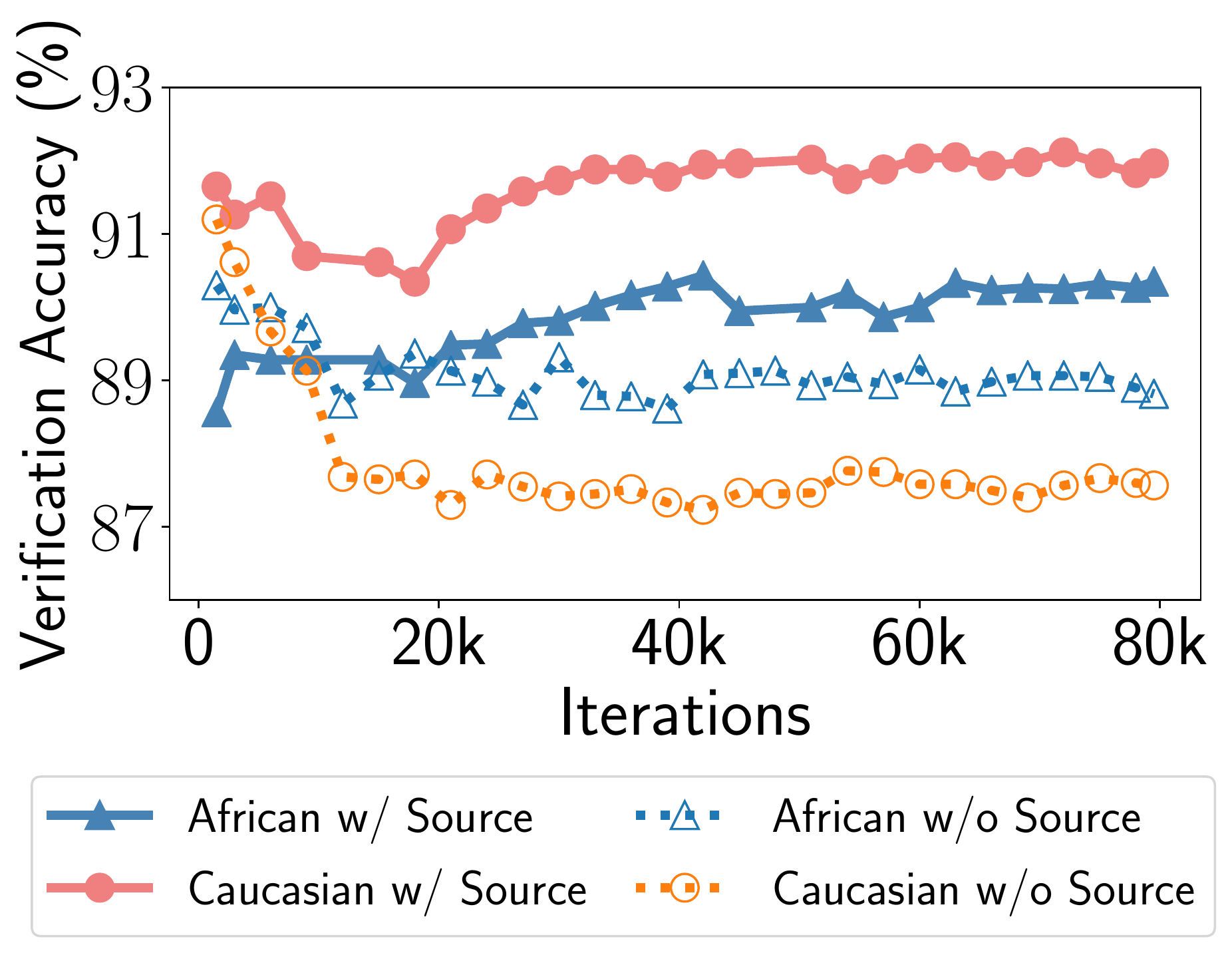}
        \caption{FedFR-L Benchmark}
    \label{fig:source-fedfr-l}
    \end{subfigure}%
    
    \caption{Performance comparison of FedFR with and without the source domain on (a) FedFR-S and (b) FedFR-L benchmark. FedFR without source domain (w/o source) hampers performances. While FedFR with source domain (w/ source) only has slight decreases on the FedFR-S and improves the performance as training proceeds on the FedFR-L.}
    \label{fig:source-domain}
\end{figure}

\textbf{Impact of Clustering Algorithm} The clustering algorithm for pseudo label generation (Section \ref{sec:clustering}) has a large impact on the overall performance of FedFR. As shown in Table \ref{tab:clusterings}, our enhanced clustering algorithm C-FINCH achieves better performance (F-score) than the other method on both FedFR-S and FedFR-L benchmarks. It leads to better performance of FedFR on the African dataset. Specifically, on the FedFR-L benchmark, C-FINCH outperforms FINCH \cite{sarfraz2019finch} by over 10\%. These experiments are run with $K = 4$. Comprehensive results of verification rate at FAR=0.1, 0.01, 0.001 are provided in the supplementary.

\textbf{Importance of Source Domain} The source domain is necessary for the federated learning process especially when its data volume is large. We compare the verification accuracy of FedFR with and without the source domain on the FedFR-S (Figure \ref{fig:source-fedfr-s}) and FedFR-L (Figure \ref{fig:source-fedfr-l}) benchmarks. FedFR without source domain (w/o source) is similar to fine-tuning. On both benchmarks, FedFR w/o source degrades dramatically on both African and Caucasian datasets. This is because of catastrophic forgetting --- the knowledge learned from the source domain data is forgotten gradually when learning in the target domain. In contrast, FedFR w/ source only has slight decreases on the FedFR-S benchmark. Moreover, its performance continues to improve as training proceeds on the FedFR-L benchmark. These results demonstrate the necessity of the source domain in the federated learning process. For a fair comparison, all experiments are conducted with $K = 4$ and without DCL.

\begingroup
\setlength{\tabcolsep}{0.35em}
\begin{table}[t]
\caption{Ablation study of FedFR on the FedFR-S and FedFR-L benchmarks.  Each added component improves the performance on the African dataset on the FedFR-L benchmark. The impact of domain constraint loss (DCL) on the FedFR-S is not as strong as on the FedFR-L because the data is balanced between domains on the FedFR-S benchmark.}
\begin{center}
\begin{tabular}{lcccccc}
\toprule
\multicolumn{1}{c}{\multirow{2}{*}{Training Method}} &
\multicolumn{5}{c}{African (\%)} &
\\
\cline{2-6}
\multicolumn{1}{c}{} &
\multicolumn{1}{c}{Acc} &
\multicolumn{1}{c}{0.1} &
\multicolumn{1}{c}{0.01} &
\multicolumn{1}{c}{0.001} &
\multicolumn{1}{c}{Rank-1} &
\\
\midrule
\textit{FedFR-S Benchmark} \\
\midrule
Pre-training (P) & 70.40 & 46.23 & 14.60 & 3.80 & 26.42 \\ 
P + Clustering (C) & 81.07 & 70.40 & 41.23 & \textbf{22.57} & 64.08 \\ 
P + C + FL & 82.18 & 74.50 & \textbf{44.23} & 20.30 & 69.42 \\ 
P + C + FL + DCL & \textbf{82.50} & \textbf{74.93} & 41.80 & 21.50 & \textbf{70.47}  \\
\midrule
\textit{FedFR-L Benchmark} \\
\midrule
Pre-training (P) & 86.65 & 83.27 & 55.93 & 37.80 & 82.13 \\ 
P + Clustering (C) & 88.72 & 87.10 & 62.23 & 35.95 & 84.05 \\ 
P + C + FL & 91.12 & 91.10 & 70.73 & 45.5 & 92.47 \\ 
P + C + FL + DCL & \textbf{91.55} & \textbf{91.57} & \textbf{76.97} & \textbf{49.40} & \textbf{93.26} \\ 
\bottomrule
\end{tabular}
\end{center}
\label{tab:ablation}
\end{table}
\endgroup

\subsection{Ablation Study}

We investigate the effects of different components in FedFR with ablation studies on the FedFR-S and FedFR-L benchmarks. We compare models obtained from four training methods, where each method contains an additional component based on the previous method: (1) \textit{Pre-training (P)}: supervised training on the source domain data. (2) \textit{P + Clustering (C)}: fine-tuning the pre-trained model on the target domain data with pseudo labels predicted by the clustering algorithm. (3) \textit{P + C + FL}: conducting federated learning based on the pre-trained model and pseudo labels for the unlabeled target domain. (4) \textit{P + C + FL + DCL}: conducting federated learning similar to (3) but with the proposed domain constraint loss (DCL) in the source domain training.

Table \ref{tab:ablation} shows the performance comparison of these four methods. Generally, the performance on the target domain (African) improves with each added component and reaches the peak with all components. Although DCL does result in superior performance under TAR@FAR=0.01 and 0.001 on the FedFR-S benchmark, it leads to the best performance on other evaluations on FedFR-S and all evaluations on FedFR-L. These results align with our intuition of DCL. We design DCL to tackle the unbalanced data problem between two domains. Since both domains on the FedFR-S benchmark contain equal data volume, it is expected that the effect of DCL on FedFR-S is not very obvious. Most importantly, DCL achieves the best results on the FedFR-L where the source domain contains more than 10x data than the target domain. Due to space constrains, we provide ablation studies on standard face recognition datasets (LFW \cite{LFWTech} and IJB-A \cite{klare2015ijb-a}) in the supplementary.

\section{Conclusion}

In this paper, we propose a novel federated unsupervised domain adaptation approach for face recognition, \textit{FedFR}, to address unsupervised domain adaptation under privacy constraints. FedFR first enhances a hierarchical clustering algorithm to generate pseudo labels for the target domain using the model pre-trained in the source domain. Then, FedFR conducts federated learning across domains using labeled data in the source domain and unlabeled data with pseudo labels in the target domain.  Additionally, we propose DCL to regularize the source domain training in federated learning, further elevating the performance in the target domain, especially when the target domain contains much less data. Extensive experiments on two constructed benchmarks demonstrate the effectiveness and significance of FedFR. We hope that methods proposed in FedFR will be useful for unsupervised domain adaptation of other multimedia applications.

\bibliographystyle{ACM-Reference-Format}
\bibliography{references}

\appendix

\end{document}